%% file: compositional_learning.tex
\documentclass{article}

\usepackage[final, nonatbib]{compositional_learning}

\usepackage[utf8]{inputenc} 
\usepackage[T1]{fontenc}    
\usepackage{hyperref}       
\usepackage{url}            
\usepackage{booktabs}       
\usepackage{amsfonts}       
\usepackage{nicefrac}       
\usepackage{microtype}      
\usepackage{xcolor}         
\usepackage{amsmath}
\usepackage{amssymb}
\usepackage{multirow}
\usepackage{makecell}
\usepackage{svg}
\usepackage{algorithm}
\usepackage{algpseudocode}

\title{Diffusion Beats Autoregressive: An Evaluation of Compositional Generation in Text-to-Image Models}

\author{%
  Arash Marioriyad \\
  Department of Computer Engineering\\
  Sharif University of Technology\\
  \texttt{arash.marioriyad98@sharif.edu} \\
  \And
  Parham Rezaei \\
  Department of Computer Engineering\\
  Sharif University of Technology\\
  \texttt{parham.rezaei@sharif.edu} \\
  \And
  Mahdieh Soleymani Baghshah \\
  Department of Computer Engineering\\
  Sharif University of Technology\\
  \texttt{soleymani@sharif.edu} \\
  \And
 Mohammad Hossein Rohban \\
  Department of Computer Engineering\\
  Sharif University of Technology\\
  \texttt{rohban@sharif.edu} \\
}

\begin{document}

\maketitle

\input{Sections/Abstract}

\input{Sections/Introduction}

\input{Sections/Models}

\input{Sections/Benchmark}

\input{Sections/Results}

\input{Sections/Conclusion}

\clearpage
\bibliography{compositional_learning}

\end{document}

%% file: Sections/Abstract.tex
\begin{abstract}
Text-to-image (T2I) generative models, such as Stable Diffusion and DALL-E, have shown remarkable proficiency in producing high-quality, realistic, and natural images from textual descriptions. However, these models sometimes fail to accurately capture all the details specified in the input prompts, particularly concerning entities, attributes, and spatial relationships. This issue becomes more pronounced when the prompt contains novel or complex compositions, leading to what are known as compositional generation failure modes. Recently, a new open-source diffusion-based T2I model, FLUX, has been introduced, demonstrating strong performance in high-quality image generation. Additionally, autoregressive T2I models like LlamaGen have claimed competitive visual quality performance compared to diffusion-based models. In this study, we evaluate the compositional generation capabilities of these newly introduced models against established models using the T2I-CompBench benchmark. Our findings reveal that LlamaGen, as a vanilla autoregressive model, is not yet on par with state-of-the-art diffusion models for compositional generation tasks under the same criteria, such as model size and inference time. On the other hand, the open-source diffusion-based model FLUX exhibits compositional generation capabilities comparable to the state-of-the-art closed-source model DALL-E3.
\end{abstract}

%% file: Sections/Introduction.tex
\section{Introduction}

Recent advancements in computational resources and data scaling have led to the development of substantial text-to-image (T2I) models, from diffusion-based \cite{ho2020denoising} models such as Stable Diffusion \cite{rombach2022high, SDXL} and DALL-E \cite{ramesh2021zero, ramesh2022hierarchical, betker2023improving} to autoregressive-based ones such as LlamaGen \cite{sun2024autoregressive}, which are capable of producing high-quality and realistic images from textual prompts. Despite these advancements, these models occasionally face difficulties in generating images that fully align with the input prompts, especially when the prompts involve complex and novel combinations of entities, attributes, and spatial relationships \cite{huang2023t2i, bakr2023hrs, li2024genai}. This challenge, known as visual compositional generation, remains a significant issue in the field of T2I generation. 

Compositional generation failure modes can be categorized into four main types: \textit{Entity missing}, \textit{incorrect attribute binding}, \textit{incorrect spatial relationship}, and \textit{incorrect numeracy}. Entity missing \cite{zhang2024enhancing, chefer2023attend, agarwal2023star, sueyoshi2024predicated, wu2024towards} is a key failure mode in T2I models, where the model omits one or more entities described in the input prompt, particularly in complex scenes involving multiple entities. Moreover, incorrect attribute binding \cite{rassin2024linguistic, feng2023trainingfree, li2023divide, wang2024compositional} occurs when an attribute, such as color, shape, size, or texture, is not faithfully bound or associated with the corresponding entity. Furthermore, incorrect spatial relationship \cite{Gokhale2022BenchmarkingSR, chatterjee2024getting} is related to the scenario where the T2I model fails to accurately capture the relative positions, orientations, or interactions between entities, resulting in a misrepresentation of the spatial arrangement described in the prompt. Finally, incorrect numeracy \cite{zafar2024iterative, kang2023counting} occurs when the model can not accurately represent the number of entities described in the input prompt, which reflects the model's limited reasoning abilities, as it struggles to maintain numerical consistency in complex scenes.

Several studies have explored the compositional generation capabilities of both diffusion \cite{okawa2024compositional} and autoregressive models \cite{ramesh2024compositional, dziri2024faith}. However, to the best of our knowledge, the field lacks a comprehensive comparison of these two generative approaches in the context of visual compositional generation from textual prompts. This study extensively evaluates the compositional generation abilities of nine state-of-the-art T2I models, including seven diffusion-based and two autoregressive-based models, using the established benchmark, T2I-CompBench \cite{huang2023t2i}.

Our results indicate that the vanilla autoregressive-based T2I model, LlamaGen \cite{sun2024autoregressive}, underperforms in all compositional generation assessments compared to SD-v1.4, the diffusion-based model most similar to LlamaGen in terms of model size (number of parameters) and inference time. This finding may suggest that adhering solely to the next-token prediction paradigm, without incorporating additional inductive biases, is insufficient to match the performance of diffusion-based approaches in compositional generation. Furthermore, an evaluation of the newly introduced open-source diffusion-based model, FLUX \cite{blackforestlab2024flux}, demonstrates that it performs competitively with the state-of-the-art closed-source T2I model, DALL-E3 \cite{betker2023improving}.

%% file: Sections/Models.tex
\section{Text-to-image Models}

This assessment evaluated nine famous T2I backbones, comprising seven diffusion-based and two autoregressive-based models (Table \ref{table:models}). These models can be categorized into five distinct families.

\paragraph{Stable Diffusion:}
Stable Diffusion models are among the most prominent open-source T2I models, utilizing a latent diffusion framework combined with an attention mechanism to process textual prompts. Specifically, the process begins with pure noise sampled from a Gaussian distribution as the initial latent code. The model then iteratively refines the latent code at each denoising step using a U-Net \cite{ronneberger2015u} architecture, which incorporates cross-attention layers to align the image generation process with the textual embedding obtained from a CLIP-based \cite{radford2021learning} model. After a predefined number of denoising steps, the final refined latent code is passed through a pre-trained image decoder to generate the final image. Through this work, we employed SD-v1.4 ,  SD-v2 \cite{rombach2022high}, and SD-XL \cite{SDXL} from the Stable Diffusion family.

\paragraph{DALL-E:}
DALL-E models are a family of closed-source diffusion-based T2I models developed and maintained by OpenAI \cite{openai2024}. While DALL-E1 \cite{ramesh2021zero} utilizes a discrete variational auto-encoder \cite{vahdat2018dvae} model to generate image tokens from textual tokens, DALL-E2 \cite{ramesh2022hierarchical} first uses a pre-trained CLIP-based model to prepare the text embeddings from the input prompt, which is then fed to a diffusion or autoregressive model to produce an image embedding. Finally, a diffusion decoder conditioned on the obtained embedding produces the final image. To further improve the prompt following abilities and image quality, DALL-E3 \cite{betker2023improving} adopts a recaptioning process of the training dataset, which is then used as the new training data for the T2I model.

\paragraph{Pixart-\text{$\boldsymbol\alpha$}:}
Pixart-\text{$\alpha$} \cite{chen2023pixart} utilizes the Diffusion Transformer (DiT) \cite{peebles2023scalable} as its core architecture, prioritizing rapid and cost-effective training compared to Stable Diffusion models. Specifically, Pixart-\text{$\alpha$} employs a three-stage training strategy, along with a recaptioning process for the training data. In the first stage, known as pixel dependency learning, Pixart-\text{$\alpha$} benefits from parameter initialization derived from an ImageNet-pretrained model and a class-guided approach to image generation. This phase focuses on generating semantically coherent pixels during a relatively inexpensive training process. The second stage involves text-image alignment learning, where Pixart-\text{$\alpha$} constructs a dataset with precise text-image pairs that exhibit high concept density, utilizing the advanced vision-language model LLaVA \cite{liu2024visual} applied to the SA-1B dataset \cite{kirillov2023segment}. In the final stage, the model is fine-tuned with high-quality aesthetic data to enhance its capability for high-resolution image generation.

\paragraph{FLUX:}
FLUX family models \cite{blackforestlab2024flux} are newly introduced T2I models developed by Black Forest Lab \cite{blackforestlab2024blog}. To the best of our knowledge, no formal technical report is available about this model. However, based on the available implementation details, the FLUX family models employ a hybrid architecture that integrates multi-modal \cite{esser2024scaling} and parallel \cite{dehghani2023scaling} diffusion transformer \cite{peebles2023scalable} blocks, operating within a flow-matching \cite{lipman2022flow} framework. Additionally, FLUX utilizes rotary positional embeddings \cite{su2024roformer} and parallel attention layers \cite{dehghani2023scaling}. The FLUX models are available in three versions: Pro, Schnell, and Dev, with the latter two being utilized in our evaluations.

\paragraph{Autoregressive:}
Vanilla Autoregressive T2I models such as LlamaGen \cite{sun2024autoregressive} employ the next-token prediction paradigm, commonly seen in large language models (LLMs), for image generation. Particularly, LlamaGen utilizes the Llama \cite{touvron2023llama} architecture for pixel generation and a quantized-autoencoder \cite{esser2021taming} framework for image tokenization. LlamaGen is introduced in two variants: class-conditioned and text-conditioned, with the latter being used for our compositional generation evaluation. For the text-conditioned variant, the model follows a two-stage training strategy. The first phase involves training on a 50-million subset of the LAION-COCO dataset \cite{laioncoco2024}at a resolution of 256×256, followed by a fine-tuning phase on 10 million internally curated high-aesthetic-quality images at a resolution of 512×512, as the second phase. LlamaGen does not incorporate a diffusion-based process, positioning it as a vanilla autoregressive T2I model without additional inductive biases.



\begin{table}[!ht]
    \centering
    \caption{Detailed Information on State-of-the-art Text-to-image Models}
    \resizebox{\textwidth}{!}{
    \begin{tabular}{cccccc}
    \cmidrule[1.5pt]{1-6}
    Model & Release Date & \#Parameters & Resolution & Training Data & Text Encoder\\
    \cmidrule[1.5pt]{1-6}
    SD-v1.4 & Aug 2022 & $860\times10^6$ & $512\times512$ & LAION-5B & ViT-L/14 CLIP\\ 
    SD-v2 & Nov 2022 & $860\times10^6$ & $768\times768$ & LAION-5B & ViT-H/14 CLIP\\ 
    SD-XL & July 2023 & $3.5 \times10^9$ & $1024\times1024$ & LAION-5B & OpenCLIP-ViT/G\\
    \cmidrule[0.8pt]{1-6}
    DALL-E3 & Nov 2023 & - & $1024\times1024$ & - & CLIP-based\\
    \cmidrule[0.8pt]{1-6}
    Pixart-\text{$\alpha$} & Sep 2023 & $600\times10^6$ & $1024\times1024$ & SA-1B & T5\\
    \cmidrule[0.8pt]{1-6}
    FLUX-Dev & Aug 2024 & - & $512\times512$ & - & T5 \& CLIP-based\\
    FLUX-Schnell & Aug 2024 & - & $1024\times1024$ & - & T5 \& CLIP-based\\
    \cmidrule[0.8pt]{1-6}
    LlamaGen-Stage1 & June 2024 & $775\times10^6$ & $256\times256$ & LAION-COCO & T5\\
    LlamaGen-Stage2 & June 2024 & $775\times10^6$ & $512\times512$ & - & T5\\
    \cmidrule[1.5pt]{1-6}
    \end{tabular}}
    \label{table:models}
\end{table}

%% file: Sections/Benchmark.tex
\section{T2I-CompBench Benchmark}

\subsection{Evaluation Datasets}
The T2I-CompBench dataset \cite{huang2023t2i} evaluates four main aspects of compositional generation capabilities: attribute binding, object relationships, numeracy, and complex compositions.

\paragraph{Attribute Binding:} This section is divided into three categories—color, shape, and texture—each comprising 300 validation prompts.

\paragraph{Object Relationships:} This part is further split into spatial and non-spatial relationships. The spatial relationships involve two sets of prompts, 2D and 3D, with 300 validation prompts each. The non-spatial relationships focus on interactions between objects and also contain 300 validation prompts.

\paragraph{Numeracy:} This section includes 300 validation prompts to assess the ability of the T2I model to understand and reason about numerical concepts.

\paragraph{Complex Compositions:} This part presents 300 validation prompts that feature more natural and challenging combinations of objects, attributes, and relationships.

\subsection{Evaluation Metrics}
T2I-CompBench \cite{huang2023t2i} employs a visual question answering (VQA) model, called BLIP-VQA \cite{li2022blip}, to evaluate the attribute binding capabilities of T2I models. For assessing spatial relationships and numeracy, the framework uses an object detector, UniDet \cite{zhou2022simple}, to estimate the relational positions and the number of objects in the generated images. For non-spatial relationships and complex compositions, T2I-CompBench utilizes several evaluation metrics: CLIP similarity score \cite{hessel2021clipscore}, which measures cosine similarity between the embeddings of the prompt and the generated image using a CLIP-based \cite{radford2021learning} model; multi-modal evaluation via a GPT-based model \cite{achiam2023gpt}; chain-of-thought prompting \cite{wei2022chain} using ShareGPT-4v \cite{chen2023sharegpt4v}; and finally, the 3-in-1 evaluation metric, which integrates CLIP similarity, BLIP-VQA, and UniDet scores.

%% file: Sections/Results.tex
\section{Results}

Table \ref{table:results_1} presents the results of attribute binding, spatial relationship, and numeracy assessments for nine state-of-the-art T2I models. DALL-E 3 and FLUX-based models demonstrate competitive performance across all aspects of compositional generation, consistently ranking at the top. Following these models, the Pixart model outperforms SD-XL in most evaluations. In contrast, the vanilla autoregressive model, LlamaGen, underperforms even when compared to the weakest Stable Diffusion model, SD-v1.4, which is comparable to LlamaGen in terms of model size and inference time. Similar trends are observed in Table \ref{table:resutls_2}, which reports the results of non-spatial relationships and complex composition assessments.

\begin{table}[!ht]
    \centering
    \caption{Quantitative Results of T2I Models on Attribute Binding, Spatial Relationship, and Numeracy}
    \resizebox{\textwidth}{!}{
    \begin{tabular}{ccccccc}
    \cmidrule[1.5pt]{1-7}
    Model & Color ($\uparrow$) & Shape ($\uparrow$) & Texture ($\uparrow$) & 2D-Spatial ($\uparrow$) & 3D-Spatial ($\uparrow$) & Numeracy ($\uparrow$)\\
    \cmidrule[1.5pt]{1-7}
    SD-v1.4 & 0.376 & 0.358 & 0.416 & 0.125 & 0.303 & 0.446\\ 
    SD-v2 & 0.506 &	0.422 & 0.492& 0.134& 0.323& 0.458 \\ 
    SD-XL & 0.588 & 0.469 & 0.530& 0.213& 0.357 & 0.499\\
    \cmidrule[0.8pt]{1-7}
    DALL-E3 & \textbf{0.778} & \textbf{0.620}& \textbf{0.704}& \underline{0.286} &	0.374 & 0.588 \\
    \cmidrule[0.8pt]{1-7}
    Pixart-\text{$\alpha$} & 0.670 & 0.493 & 0.648 & 0.206	& \underline{0.390}& 0.506\\
    \cmidrule[0.8pt]{1-7}
    FLUX-Dev & \underline{0.771}&	0.495&	0.604 &	0.266 &	0.384 &	\textbf{0.618}\\
    FLUX-Schnell & 0.740 &	\underline{0.571} &	\underline{0.685}&	\textbf{0.292}&	\textbf{0.391}&	\underline{0.606}\\
    \cmidrule[0.8pt]{1-7}
    LlamaGen-Stage1 & 0.271 &	0.391 &	0.492 &	0.084&	0.227 & 0.357\\
    LlamaGen-Stage2 & 0.285	& 0.329 &0.373 &	0.119 &	0.155 &	0.265\\
    \cmidrule[1.5pt]{1-7}
    \end{tabular}}
    \label{table:results_1}
\end{table}

\begin{table}[!ht]
    \centering
    \caption{Quantitative Results of T2I Models on Non-spatial Relationship and Complex Compositions}
    \resizebox{\textwidth}{!}{
    \begin{tabular}{clcccccccccc}
    
    \cmidrule[1.5pt]{2-9}    
    
    && \multicolumn{3}{c}{Non-spatial ($\uparrow$)} 
    && \multicolumn{3}{c}{Complex ($\uparrow$)} \\
    
    \cmidrule[1.25pt]{3-5}
    \cmidrule[1.25pt]{7-9}
    
    & Models & CLIP & GPT-4v  & Share-CoT  && 
    3-in-1  & GPT-4v & Share-CoT \\
    
    \cmidrule[1.5pt]{2-9}
     & SD-v1.4 & 0.308 & 0.820 & 0.749 & & 0.308	& 0.714& 0.773 \\
     & SD-XL & \underline{0.312}& 0.844 & 0.767  & & 0.324 & 0.756 & 0.782 \\ 
    \cmidrule[0.8pt]{2-9}
    & DALL-E3 & 0.300 & \textbf{0.927} & \textbf{0.793} && \textbf{0.377} & \textbf{0.828}&	\textbf{0.793}\\
    \cmidrule[0.8pt]{2-9}
    & FLUX-Dev & 0.306 & \underline{0.874}& 0.780 && 0.364 &	0.794&\underline{0.791}\\
    & FLUX-Schnell & \textbf{0.313} & 0.872 & \underline{0.784}& & \underline{0.368} &	\underline{0.823}&\textbf{0.793}\\
    \cmidrule[0.8pt]{2-9}
    & LlamaGen-Stage1 & 0.305 &	0.788 & 0.783 &&	0.283 & 	0.584 & 0.769\\
    & LlamaGen-Stage2 & 0.272 &	0.669 & 0.763 & &0.255 &	0.562 & 0.765\\
    \cmidrule[1.5pt]{2-9}
    \end{tabular}}
    \label{table:resutls_2}
\end{table}

%% file: Sections/Conclusion.tex
\section{Discussion}

These findings indicate that the pure next-token prediction paradigm fails to compete effectively with diffusion-based models of similar size in the absence of inductive biases tailored to the visual generation domain. Notably, While the class-conditioned version of LlamaGen exhibits competitive performance in terms of image quality and naturalness compared to diffusion models, the weaker performance of the text-conditioned version suggests that autoregressive models may face more significant challenges in capturing complex conditions. Furthermore, considering the critical role of tokenization in autoregressive models, selecting an appropriate image tokenizer with suitable granularity may enhance LlamaGen's compositional generation capabilities. Ultimately, the inductive bias of the mask image modeling or next-token prediction paradigm may not be sufficient for generating images that are fully aligned with the textual prompts.

%% file: compositional_learning.bbl
\begin{thebibliography}{10}

\bibitem{achiam2023gpt}
Josh Achiam, Steven Adler, Sandhini Agarwal, Lama Ahmad, Ilge Akkaya, Florencia~Leoni Aleman, Diogo Almeida, Janko Altenschmidt, Sam Altman, Shyamal Anadkat, et~al.
\newblock Gpt-4 technical report.
\newblock {\em arXiv preprint arXiv:2303.08774}, 2023.

\bibitem{agarwal2023star}
Aishwarya Agarwal, Srikrishna Karanam, KJ~Joseph, Apoorv Saxena, Koustava Goswami, and Balaji~Vasan Srinivasan.
\newblock A-star: Test-time attention segregation and retention for text-to-image synthesis.
\newblock In {\em Proceedings of the IEEE/CVF International Conference on Computer Vision}, pages 2283--2293, 2023.

\bibitem{bakr2023hrs}
Eslam~Mohamed Bakr, Pengzhan Sun, Xiaogian Shen, Faizan~Farooq Khan, Li~Erran Li, and Mohamed Elhoseiny.
\newblock Hrs-bench: Holistic, reliable and scalable benchmark for text-to-image models.
\newblock In {\em Proceedings of the IEEE/CVF International Conference on Computer Vision}, pages 20041--20053, 2023.

\bibitem{betker2023improving}
James Betker, Gabriel Goh, Li~Jing, Tim Brooks, Jianfeng Wang, Linjie Li, Long Ouyang, Juntang Zhuang, Joyce Lee, Yufei Guo, et~al.
\newblock Improving image generation with better captions.
\newblock {\em Computer Science. https://cdn. openai. com/papers/dall-e-3. pdf}, 2(3):8, 2023.

\bibitem{chatterjee2024getting}
Agneet Chatterjee, Gabriela Ben~Melech Stan, Estelle Aflalo, Sayak Paul, Dhruba Ghosh, Tejas Gokhale, Ludwig Schmidt, Hannaneh Hajishirzi, Vasudev Lal, Chitta Baral, et~al.
\newblock Getting it right: Improving spatial consistency in text-to-image models.
\newblock {\em arXiv preprint arXiv:2404.01197}, 2024.

\bibitem{chefer2023attend}
Hila Chefer, Yuval Alaluf, Yael Vinker, Lior Wolf, and Daniel Cohen-Or.
\newblock Attend-and-excite: Attention-based semantic guidance for text-to-image diffusion models.
\newblock {\em ACM Transactions on Graphics (TOG)}, 42(4):1--10, 2023.

\bibitem{chen2023pixart}
Junsong Chen, Jincheng Yu, Chongjian Ge, Lewei Yao, Enze Xie, Yue Wu, Zhongdao Wang, James Kwok, Ping Luo, Huchuan Lu, et~al.
\newblock Pixart-$\backslash \alpha $: Fast training of diffusion transformer for photorealistic text-to-image synthesis.
\newblock {\em arXiv preprint arXiv:2310.00426}, 2023.

\bibitem{chen2023sharegpt4v}
Lin Chen, Jisong Li, Xiaoyi Dong, Pan Zhang, Conghui He, Jiaqi Wang, Feng Zhao, and Dahua Lin.
\newblock Sharegpt4v: Improving large multi-modal models with better captions.
\newblock {\em arXiv preprint arXiv:2311.12793}, 2023.

\bibitem{dehghani2023scaling}
Mostafa Dehghani, Josip Djolonga, Basil Mustafa, Piotr Padlewski, Jonathan Heek, Justin Gilmer, Andreas~Peter Steiner, Mathilde Caron, Robert Geirhos, Ibrahim Alabdulmohsin, et~al.
\newblock Scaling vision transformers to 22 billion parameters.
\newblock In {\em International Conference on Machine Learning}, pages 7480--7512. PMLR, 2023.

\bibitem{dziri2024faith}
Nouha Dziri, Ximing Lu, Melanie Sclar, Xiang~Lorraine Li, Liwei Jiang, Bill~Yuchen Lin, Sean Welleck, Peter West, Chandra Bhagavatula, Ronan Le~Bras, et~al.
\newblock Faith and fate: Limits of transformers on compositionality.
\newblock {\em Advances in Neural Information Processing Systems}, 36, 2024.

\bibitem{esser2024scaling}
Patrick Esser, Sumith Kulal, Andreas Blattmann, Rahim Entezari, Jonas M{\"u}ller, Harry Saini, Yam Levi, Dominik Lorenz, Axel Sauer, Frederic Boesel, et~al.
\newblock Scaling rectified flow transformers for high-resolution image synthesis.
\newblock In {\em Forty-first International Conference on Machine Learning}, 2024.

\bibitem{esser2021taming}
Patrick Esser, Robin Rombach, and Bjorn Ommer.
\newblock Taming transformers for high-resolution image synthesis.
\newblock In {\em Proceedings of the IEEE/CVF conference on computer vision and pattern recognition}, pages 12873--12883, 2021.

\bibitem{feng2023trainingfree}
Weixi Feng, Xuehai He, Tsu-Jui Fu, Varun Jampani, Arjun~Reddy Akula, Pradyumna Narayana, Sugato Basu, Xin~Eric Wang, and William~Yang Wang.
\newblock Training-free structured diffusion guidance for compositional text-to-image synthesis.
\newblock In {\em The Eleventh International Conference on Learning Representations}, 2023.

\bibitem{Gokhale2022BenchmarkingSR}
Tejas Gokhale, Hamid Palangi, Besmira Nushi, Vibhav Vineet, Eric Horvitz, Ece Kamar, Chitta Baral, and Yezhou Yang.
\newblock Benchmarking spatial relationships in text-to-image generation.
\newblock {\em ArXiv}, abs/2212.10015, 2022.

\bibitem{hessel2021clipscore}
Jack Hessel, Ari Holtzman, Maxwell Forbes, Ronan~Le Bras, and Yejin Choi.
\newblock Clipscore: A reference-free evaluation metric for image captioning.
\newblock {\em arXiv preprint arXiv:2104.08718}, 2021.

\bibitem{ho2020denoising}
Jonathan Ho, Ajay Jain, and Pieter Abbeel.
\newblock Denoising diffusion probabilistic models.
\newblock {\em Advances in neural information processing systems}, 33:6840--6851, 2020.

\bibitem{huang2023t2i}
Kaiyi Huang, Kaiyue Sun, Enze Xie, Zhenguo Li, and Xihui Liu.
\newblock T2i-compbench: A comprehensive benchmark for open-world compositional text-to-image generation.
\newblock {\em Advances in Neural Information Processing Systems}, 36:78723--78747, 2023.

\bibitem{kang2023counting}
Wonjun Kang, Kevin Galim, and Hyung~Il Koo.
\newblock Counting guidance for high fidelity text-to-image synthesis.
\newblock {\em arXiv preprint arXiv:2306.17567}, 2023.

\bibitem{kirillov2023segment}
Alexander Kirillov, Eric Mintun, Nikhila Ravi, Hanzi Mao, Chloe Rolland, Laura Gustafson, Tete Xiao, Spencer Whitehead, Alexander~C Berg, Wan-Yen Lo, et~al.
\newblock Segment anything.
\newblock In {\em Proceedings of the IEEE/CVF International Conference on Computer Vision}, pages 4015--4026, 2023.

\bibitem{blackforestlab2024blog}
Black~Forest Lab.
\newblock \url{https://www.blackforestlab.com/blog}, 2024.
\newblock Accessed: 2024-09.

\bibitem{blackforestlab2024flux}
Black~Forest Lab.
\newblock Flux: A diffusion-based text-to-image (t2i) model.
\newblock \url{https://github.com/blackforestlab/flux}, 2024.
\newblock Accessed: 2024-09.

\bibitem{laioncoco2024}
LAION.
\newblock Laion-coco 600m.
\newblock \url{https://laion.ai/blog/laion-coco}, 2022.

\bibitem{li2024genai}
Baiqi Li, Zhiqiu Lin, Deepak Pathak, Jiayao Li, Yixin Fei, Kewen Wu, Tiffany Ling, Xide Xia, Pengchuan Zhang, Graham Neubig, et~al.
\newblock Genai-bench: Evaluating and improving compositional text-to-visual generation.
\newblock {\em arXiv preprint arXiv:2406.13743}, 2024.

\bibitem{li2022blip}
Junnan Li, Dongxu Li, Caiming Xiong, and Steven Hoi.
\newblock Blip: Bootstrapping language-image pre-training for unified vision-language understanding and generation.
\newblock In {\em International conference on machine learning}, pages 12888--12900. PMLR, 2022.

\bibitem{li2023divide}
Yumeng Li, Margret Keuper, Dan Zhang, and Anna Khoreva.
\newblock Divide \& bind your attention for improved generative semantic nursing.
\newblock In {\em 34th British Machine Vision Conference 2023, {BMVC} 2023}, 2023.

\bibitem{lipman2022flow}
Yaron Lipman, Ricky~TQ Chen, Heli Ben-Hamu, Maximilian Nickel, and Matt Le.
\newblock Flow matching for generative modeling.
\newblock {\em arXiv preprint arXiv:2210.02747}, 2022.

\bibitem{liu2024visual}
Haotian Liu, Chunyuan Li, Qingyang Wu, and Yong~Jae Lee.
\newblock Visual instruction tuning.
\newblock {\em Advances in neural information processing systems}, 36, 2024.

\bibitem{okawa2024compositional}
Maya Okawa, Ekdeep~S Lubana, Robert Dick, and Hidenori Tanaka.
\newblock Compositional abilities emerge multiplicatively: Exploring diffusion models on a synthetic task.
\newblock {\em Advances in Neural Information Processing Systems}, 36, 2024.

\bibitem{openai2024}
OpenAI.
\newblock \url{https://www.openai.com}, 2024.
\newblock Accessed: 2024-09.

\bibitem{peebles2023scalable}
William Peebles and Saining Xie.
\newblock Scalable diffusion models with transformers.
\newblock In {\em Proceedings of the IEEE/CVF International Conference on Computer Vision}, pages 4195--4205, 2023.

\bibitem{SDXL}
Dustin Podell, Zion English, Kyle Lacey, Andreas Blattmann, Tim Dockhorn, Jonas Müller, Joe Penna, and Robin Rombach.
\newblock Sdxl: Improving latent diffusion models for high-resolution image synthesis, 2023.

\bibitem{radford2021learning}
Alec Radford, Jong~Wook Kim, Chris Hallacy, Aditya Ramesh, Gabriel Goh, Sandhini Agarwal, Girish Sastry, Amanda Askell, Pamela Mishkin, Jack Clark, et~al.
\newblock Learning transferable visual models from natural language supervision.
\newblock In {\em International conference on machine learning}, pages 8748--8763. PMLR, 2021.

\bibitem{ramesh2022hierarchical}
Aditya Ramesh, Prafulla Dhariwal, Alex Nichol, Casey Chu, and Mark Chen.
\newblock Hierarchical text-conditional image generation with clip latents.
\newblock {\em arXiv preprint arXiv:2204.06125}, 1(2):3, 2022.

\bibitem{ramesh2021zero}
Aditya Ramesh, Mikhail Pavlov, Gabriel Goh, Scott Gray, Chelsea Voss, Alec Radford, Mark Chen, and Ilya Sutskever.
\newblock Zero-shot text-to-image generation.
\newblock In {\em International conference on machine learning}, pages 8821--8831. Pmlr, 2021.

\bibitem{ramesh2024compositional}
Rahul Ramesh, Ekdeep~Singh Lubana, Mikail Khona, Robert~P. Dick, and Hidenori Tanaka.
\newblock Compositional capabilities of autoregressive transformers: A study on synthetic, interpretable tasks.
\newblock In {\em Forty-first International Conference on Machine Learning}, 2024.

\bibitem{rassin2024linguistic}
Royi Rassin, Eran Hirsch, Daniel Glickman, Shauli Ravfogel, Yoav Goldberg, and Gal Chechik.
\newblock Linguistic binding in diffusion models: Enhancing attribute correspondence through attention map alignment.
\newblock {\em Advances in Neural Information Processing Systems}, 36, 2024.

\bibitem{rombach2022high}
Robin Rombach, Andreas Blattmann, Dominik Lorenz, Patrick Esser, and Bj{\"o}rn Ommer.
\newblock High-resolution image synthesis with latent diffusion models.
\newblock In {\em Proceedings of the IEEE/CVF conference on computer vision and pattern recognition}, pages 10684--10695, 2022.

\bibitem{ronneberger2015u}
Olaf Ronneberger, Philipp Fischer, and Thomas Brox.
\newblock U-net: Convolutional networks for biomedical image segmentation.
\newblock In {\em Medical image computing and computer-assisted intervention--MICCAI 2015: 18th international conference, Munich, Germany, October 5-9, 2015, proceedings, part III 18}, pages 234--241. Springer, 2015.

\bibitem{su2024roformer}
Jianlin Su, Murtadha Ahmed, Yu~Lu, Shengfeng Pan, Wen Bo, and Yunfeng Liu.
\newblock Roformer: Enhanced transformer with rotary position embedding.
\newblock {\em Neurocomputing}, 568:127063, 2024.

\bibitem{sueyoshi2024predicated}
Kota Sueyoshi and Takashi Matsubara.
\newblock Predicated diffusion: Predicate logic-based attention guidance for text-to-image diffusion models.
\newblock In {\em Proceedings of the IEEE/CVF Conference on Computer Vision and Pattern Recognition}, pages 8651--8660, 2024.

\bibitem{sun2024autoregressive}
Peize Sun, Yi~Jiang, Shoufa Chen, Shilong Zhang, Bingyue Peng, Ping Luo, and Zehuan Yuan.
\newblock Autoregressive model beats diffusion: Llama for scalable image generation.
\newblock {\em arXiv preprint arXiv:2406.06525}, 2024.

\bibitem{touvron2023llama}
Hugo Touvron, Thibaut Lavril, Gautier Izacard, Xavier Martinet, Marie-Anne Lachaux, Timoth{\'e}e Lacroix, Baptiste Rozi{\`e}re, Naman Goyal, Eric Hambro, Faisal Azhar, et~al.
\newblock Llama: Open and efficient foundation language models.
\newblock {\em arXiv preprint arXiv:2302.13971}, 2023.

\bibitem{vahdat2018dvae}
Arash Vahdat, Evgeny Andriyash, and William Macready.
\newblock Dvae\#: Discrete variational autoencoders with relaxed boltzmann priors.
\newblock {\em Advances in Neural Information Processing Systems}, 31, 2018.

\bibitem{wang2024compositional}
Ruichen Wang, Zekang Chen, Chen Chen, Jian Ma, Haonan Lu, and Xiaodong Lin.
\newblock Compositional text-to-image synthesis with attention map control of diffusion models.
\newblock In {\em Proceedings of the AAAI Conference on Artificial Intelligence}, volume~38, pages 5544--5552, 2024.

\bibitem{wei2022chain}
Jason Wei, Xuezhi Wang, Dale Schuurmans, Maarten Bosma, Fei Xia, Ed~Chi, Quoc~V Le, Denny Zhou, et~al.
\newblock Chain-of-thought prompting elicits reasoning in large language models.
\newblock {\em Advances in neural information processing systems}, 35:24824--24837, 2022.

\bibitem{wu2024towards}
Yihang Wu, Xiao Cao, Kaixin Li, Zitan Chen, Haonan Wang, Lei Meng, and Zhiyong Huang.
\newblock Towards better text-to-image generation alignment via attention modulation.
\newblock {\em arXiv preprint arXiv:2404.13899}, 2024.

\bibitem{zafar2024iterative}
Oz~Zafar, Lior Wolf, and Idan Schwartz.
\newblock Iterative object count optimization for text-to-image diffusion models.
\newblock {\em arXiv preprint arXiv:2408.11721}, 2024.

\bibitem{zhang2024enhancing}
Yang Zhang, Teoh~Tze Tzun, Lim~Wei Hern, Tiviatis Sim, and Kenji Kawaguchi.
\newblock Enhancing semantic fidelity in text-to-image synthesis: Attention regulation in diffusion models.
\newblock {\em arXiv preprint arXiv:2403.06381}, 2024.

\bibitem{zhou2022simple}
Xingyi Zhou, Vladlen Koltun, and Philipp Kr{\"a}henb{\"u}hl.
\newblock Simple multi-dataset detection.
\newblock In {\em Proceedings of the IEEE/CVF conference on computer vision and pattern recognition}, pages 7571--7580, 2022.

\end{thebibliography}
